%% file: main.tex
\pdfoutput=1

\documentclass[11pt]{article}

\usepackage{ACL2023}
\usepackage{times}
\usepackage{latexsym}

\usepackage[T1]{fontenc}

\usepackage[utf8]{inputenc}

\usepackage{microtype}

\usepackage{inconsolata}
\usepackage{url}

\input{math_commands.tex}

\usepackage{amsmath}
\usepackage{multirow}
\usepackage{xcolor}

\usepackage{xcolor,colortbl}
\definecolor{lightblue}{rgb}{0.68, 0.85, 0.9}
\definecolor{lavender}{rgb}{0.9, 0.9, 0.98}
\definecolor{lightyellow}{rgb}{1.0, 1.0, 0.88}
\definecolor{magicmint}{rgb}{0.67, 0.94, 0.82}
\definecolor{palepink}{rgb}{0.98, 0.85, 0.87}
\definecolor{bubbles}{rgb}{0.91, 1.0, 1.0}

\usepackage{microtype}
\usepackage{graphicx}
\usepackage{subfigure}
\usepackage{amssymb}
\usepackage{CJKutf8}
\usepackage{algorithm}
\usepackage{algorithmic}
\usepackage{enumitem}
\algsetup{linenosize=\small}
\usepackage{wrapfig}
\usepackage{cleveref}
\usepackage{multicol}
\usepackage{booktabs}
\usepackage{xspace}
\usepackage{adjustbox}
\usepackage{subfiles}
\usepackage{latexsym}

\usepackage{placeins}
\usepackage[french,vietnamese,main=english]{babel}
\usepackage{tikz}
\babelprovide[import]{thai}

\newcommand*{\affaddr}[1]{#1} 
\newcommand*{\affmark}[1][*]{\textsuperscript{#1}}
\newcommand*{\email}[1]{\textrm{#1}}

\title{Is GPT-3 a Good Data Annotator?}

\author{Bosheng Ding\thanks{\; Equal contribution, order decided by coin flip. Bosheng Ding are under the Joint PhD Program between Alibaba and Nanyang Technological University.}\affmark[~~1,2]~~Chengwei Qin\footnotemark[1]\affmark[~~1]~~Linlin Liu\thanks{\; Work done while under the Joint PhD Program between Alibaba and Nanyang Technological University.}~\affmark[1,2]\\\textbf{Yew Ken Chia\affmark[2]}~~\textbf{
Boyang Li\affmark[1]~~}
\textbf{
Shafiq Joty\affmark[1]~~}\textbf{Lidong Bing\thanks{\; Correspondent Author.}~\affmark[2]}~~ \\
\affaddr{\affmark[1]Nanyang Technological University, Singapore}
\affaddr{\affmark[2]DAMO Academy, Alibaba Group}\\
\email{\small{\{bosheng001, chengwei003, linlin001, boyang.li, srjoty\}@ntu.edu.sg}}\\
\email{\small{\{bosheng.ding, yewken.chia, l.bing\}@alibaba-inc.com}}}

\begin{document}
\maketitle
\begin{abstract}

\input{00_abstract}
\end{abstract}

\input{01_introduction}

\input{02_related_work}

\input{03_methodology}
\input{04_experiments}

\input{05_analysis}
\input{06_conclusion}

\bibliography{acl2023}
\bibliographystyle{acl_natbib}

\appendix
\newpage
\input{07_appendix}

\end{document}

%% file: math_commands.tex
\usepackage{amsmath,amsfonts,bm}

\def\eqref#1{equation~\ref{#1}}

\def\1{\bm{1}}

\DeclareMathAlphabet{\mathsfit}{\encodingdefault}{\sfdefault}{m}{sl}
\SetMathAlphabet{\mathsfit}{bold}{\encodingdefault}{\sfdefault}{bx}{n}



%% file: 00_abstract.tex
Data annotation is the process of labeling data that could be used to train machine learning models. Having high-quality annotation is crucial, as it allows the model to learn the relationship between the input data and the desired output. GPT-3, a large-scale language model developed by OpenAI, has demonstrated impressive zero- and few-shot performance on a wide range of NLP tasks. It is therefore natural to wonder whether it can be used to effectively annotate data for NLP tasks. In this paper, we evaluate the performance of GPT-3 as a data annotator by comparing it with traditional data annotation methods and analyzing its output on a range of tasks. Through this analysis, we aim to provide insight into the potential of GPT-3 as a general-purpose data annotator in NLP 
\footnote{Our code is available at \url{https://github.com/DAMO-NLP-SG/LLM-Data-Annotator}.}.

%% file: 01_introduction.tex
\section{Introduction}
\label{sec:introduction}


The democratization of artificial intelligence (AI) \cite{Garvey2018AFF, Rubeis2022DemocratizingAI} aims to provide access to AI technologies to all members of society, including individuals, small- and medium-sized enterprises (SMEs), academic research labs, and nonprofit organizations. Achieving this goal is crucial for the promotion of innovation, economic growth, and fairness and equality. As typical AI models are usually data-hungry, one significant obstacle of AI democratization is the preparation of well-annotated data for training AI models.


Specifically, supervised learning critically depends on sufficient training data with accurate annotation, but data annotation can be a costly endeavor, particularly for small-scale companies and organizations \cite{Bunte2021WhyII}. The cost of data annotation typically includes the labor costs associated with the labeling process, as well as the time and resources required to hire, train and manage annotators. Additionally, there may be costs associated with the annotation tools and infrastructure needed to support the annotation process. Individuals or small-scale organizations may not have resources to annotate sufficient training data, thereby are unable to reap the benefits of contemporary AI technologies. 
Although the development of pre-trained language models such as BERT \cite{Devlin2019BERTPO}, XLNet \cite{Yang2019XLNetGA}, GPT-2 \cite{radford2019language} and RoBERTa \cite{Liu2019RoBERTaAR} eases the data-hungry issue to some extent, data annotation remains an unavoidable challenge for supervised model training. 


GPT-3 \cite{Brown2020LanguageMA, Ouyang2022TrainingLM}\footnote{For brevity, we refer to both the original GPT-3 and InstructGPT as GPT-3.} is a powerful large language model developed by OpenAI. 
Evaluations show that GPT-3 has gained through pretraining a surprisingly wide range of knowledge, which can be transferred to downstream tasks through knowledge distillation \cite{Kim2022AskMW}. We present some examples in Appendix~\ref{sec:specific_domain}. Due to the model architecture and pretraining tasks designed for auto-regressive generation, GPT-3 is capable of generating human-like text and performing a broad array of NLP tasks, such as machine translation, summarization, and question-answering. 
However, the direct use of GPT-3 for inference in a production setting remains challenging due to its size and computational requirements. Moreover, such large language models often lack the flexibility of local deployment, since their parameters are usually not publicly available. In contrast, it is often more feasible to use smaller language model models, such as BERT$_\mathrm{BASE}$ \cite{Devlin2019BERTPO}, in production environments.

In this paper, we investigate the ability of GPT-3 to annotate training data for training machine learning models, which can substantially lower the annotation cost and level the playing field for individuals or small organizations, so that they can harness the power of AI in their own missions. The process can be considered as distilling the knowledge of GPT-3 to small networks that can be straightforwardly deployed in production environments. 

We conduct extensive experiments to evaluate the performance, time, and cost-effectiveness of 3 different GPT-3 based data annotation approaches for both sequence- and token-level NLP tasks. Our main contributions can be summarized as follows: 

\vspace{-0.5em}
\begin{itemize} [leftmargin=10pt]\itemsep-0.2em
    \item We conduct comprehensive analysis of the feasibility of leveraging GPT-3 for data annotation for complex NLP tasks.
    \item We study 3 different GPT-3 based data annotation approaches, and then conduct extensive experiments on both sequence- and token-level NLP tasks to evaluate their performance.
    \item We find that directly annotating unlabeled data is suitable for tasks with small label space while generation-based methods are more suitable for tasks with large label space. 
    \item We find that generation-based approaches tend to be more cost-effective compared with directly annotating unlabeled data.
\end{itemize}

%% file: 02_related_work.tex
\section{Related Work}
\label{sec:related_work}
\paragraph{Large Language Models}
Large language models (LLMs) have made significant progress on natural language processing tasks in recent years. These models are trained with self-supervision on large, general corpora and demonstrate excellent performance on numerous tasks \cite{Brown2020LanguageMA, Rae2021ScalingLM, Taylor2022GalacticaAL, Hoffmann2022TrainingCL, Black2022GPTNeoX20BAO, Zhang2022OPTOP, Chowdhery2022PaLMSL,Thoppilan2022LaMDALM, touvron2023llama}. LLMs possess the ability to learn in context through few-shot learning \cite{Brown2020LanguageMA, Ouyang2022TrainingLM}. Their capabilities expand with scale, and recent research has highlighted their ability to reason at larger scales with an appropriate prompting strategy \cite{Lester2021ThePO, wei2022chain, Chowdhery2022PaLMSL, Liu2021PretrainPA, Kojima2022LargeLM, Lewkowycz2022SolvingQR,qin2023chatgpt,zhao2023retrieving,Li2023ChainOK,Jiao2023LogicLLMES}. 

\citet{Wang2021WantTR} investigate methods to utilize GPT-3 to annotate unlabeled data. However, they mainly focus on the generation and sequence classification tasks. In this work, we conduct more comprehensive experiments and analysis on a wider range of settings, covering both sequence- and token-level tasks. In a recent work, \citet{Liu2022WANLIWA} demonstrate a worker-and-AI collaborative approach for dataset creation with a few seed examples, while we also analyze approaches that support zero-shot training data generation, which do not require any seed examples.
\paragraph{Prompt-Learning}
Prompt-Learning, also known as Prompting, offers insight into what the future of NLP may look like \cite{Lester2021ThePO, Liu2021PretrainPA, Ding2021OpenPromptAO}. By mimicking the process of pre-training, prompt-learning intuitively connects pre-training and model tuning \cite{Liu2021PTuningVP}. In practice, this paradigm has proven remarkably effective in low-data regimes \cite{Scao2021HowMD,Gao2021MakingPL,qin2022lfpt}. For instance, with an appropriate template, zero-shot prompt-learning can even outperform 32-shot fine-tuning \cite{Ding2021PromptLearningFF}. Another promising characteristic of prompt-learning is its potential to stimulate large-scale pre-trained language models (PLMs). When applied to a 10B model, optimizing prompts alone (while keeping the parameters of the model fixed) can yield comparable performance to full parameter fine-tuning \cite{Lester2021ThePO,qin2023learning}. These practical studies suggest that prompts can be used to more effectively and efficiently extract knowledge from PLMs, leading to a deeper understanding of the underlying principles of their mechanisms \cite{Li2022DoesGD}.
\paragraph{Data Augmentation}
There has been a significant amount of research in NLP on learning with limited labeled data for various tasks, including unsupervised pre-training \cite{Devlin2019BERTPO, Peters2018DeepCW, Yang2019XLNetGA, Raffel2019ExploringTL, Liu2021EnhancingML}, multi-task learning \cite{Glorot2011DomainAF, Liu2017AdversarialML}, semi-supervised learning \cite{Miyato2016AdversarialTM}, and few-shot learning \cite{Deng2019WhenLR, He2021OnTE, qin-joty-2022-continual}. One approach to address the need for labeled data is through data augmentation \cite{Feng2021ASO, meng2022generating, Chen2021AnES}, which involves generating new data by modifying existing data points using transformations based on prior knowledge about the problem's structure \cite{Yang2020GenerativeDA}. The augmented data can be generated from labeled data \cite{Ding2020DAGADA, Liu2021MulDAAM, ding-etal-2022-globalwoz} and used directly in supervised learning \cite{Wei2019EDAED} or employed in semi-supervised learning for unlabeled data through consistency regularization \cite{Xie2019UnsupervisedDA}.

%% file: 03_methodology.tex
\section{Methodology}
\label{sec:framework}
\begin{figure}[t!]
    \centering
    \includegraphics[scale=0.35]{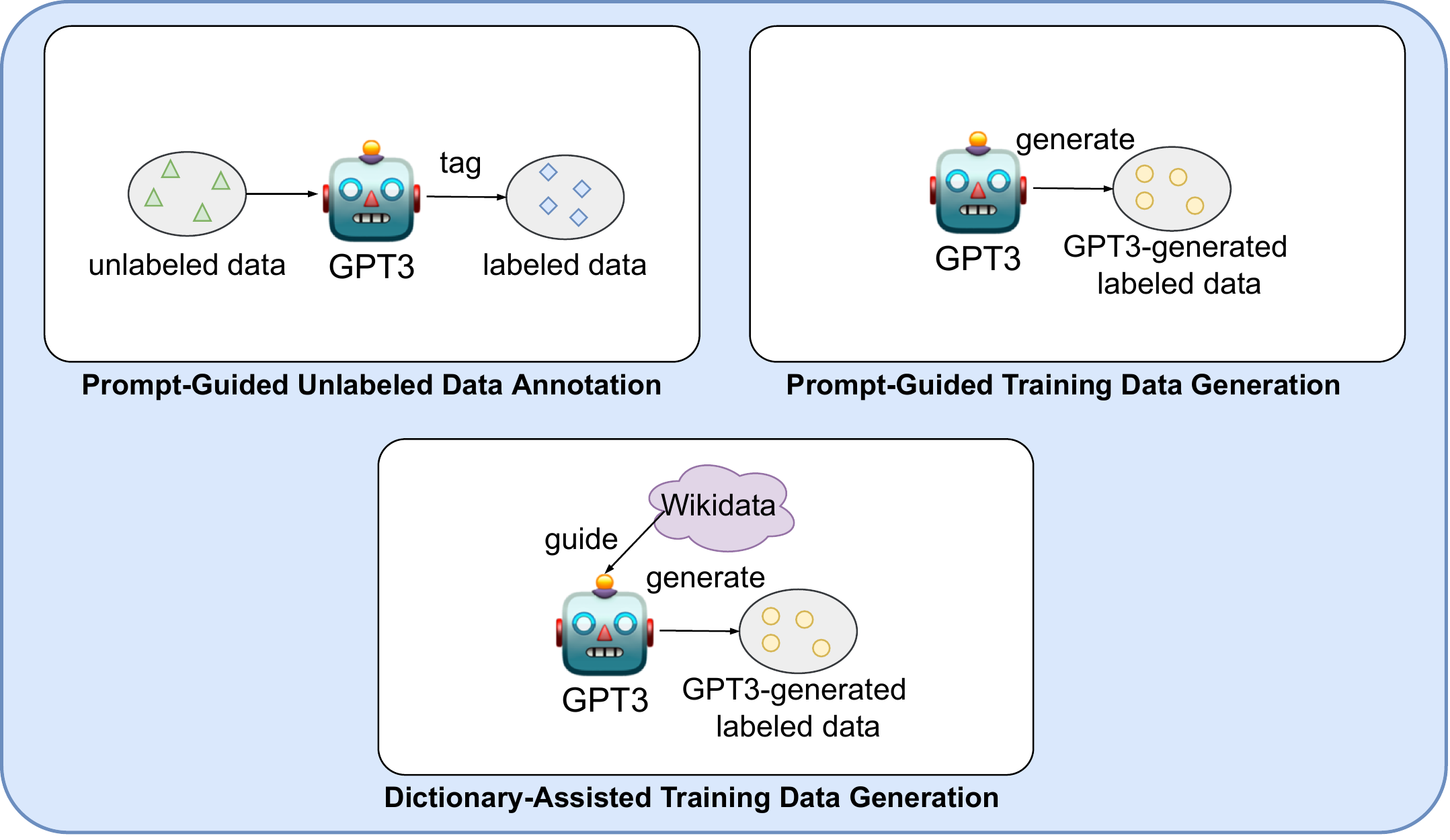}
    \vspace{-1.5em}
    \caption{Illustrations of our proposed methods.}
    \label{fig:method} 
\end{figure}

We study 3 different approaches to utilize GPT-3 for data annotation: 1) prompt-guided unlabeled data annotation (PGDA); 2) prompt-guided training data generation (PGDG); and  3) dictionary-assisted training data generation (DADG). Illustrations are shown in Figure~\ref{fig:method}. Overall, these 3 approaches can be regarded as in-context learning \cite{wei2022chain}, a new paradigm that is getting popular in NLP. Under this paradigm, a language model “learns” to do a task simply by conditioning on $l_\mathrm{IOP}$, a list of input-output pairs (IOP). \footnote{Under the zero-shot settings, where $l_\mathrm{IOP}$ is not provided, our methods become instruction-tuning \cite{wei2021finetuned}.} More formally,
\begin{align}
    y_i = \text{GPT-3} (l_\mathrm{IOP}, x_i)
\end{align}
where $x_i$ is the query input sequence and $y_i$ is the text generated by GPT-3.
For comparison, the performance, cost, and time spent on the three methods are monitored. We also report the results of \textbf{Prompted Direct Inference (PGI)}, which is to instruct GPT-3 to directly annotate the test data. 

 \subsection{Prompt-Guided Unlabeled Data Annotation (PGDA)}
 \label{sec:framework1}
The first approach involves the creation of prompts to guide GPT-3 in annotating unlabeled data. To this end, task-specific prompts are designed to elicit labels from GPT-3 for a given set of unlabeled data. In our experiments, the unlabeled data is derived from human-labeled datasets by removing the existing labels. The resulting GPT-3-labeled data is then used to train a local model to predict human-labeled test data, with the performance of this model being evaluated. As shown in Figure~\ref{fig:prompt_example_sst2}, an instruction with few-shot examples is given to GPT-3, followed by unlabeled data. GPT-3 is then prompted to predict labels for the unlabeled data. 

\begin{figure}[t!]
    \centering
    \begin{tikzpicture}
    \node[draw, rounded corners] {
    \resizebox{0.8\linewidth}{!}{    
    \adjustbox{minipage=[r][11em][b]{0.4\textwidth},scale={0.7}}{
    \textbf{Choose the sentiment of the given text from Positive and Negative.}\\
    \textbf{\textcolor{blue}{Text:}} a feast for the eyes\\
    \textbf{\textcolor{blue}{Sentiment:}} Positive\\
     \textbf{\textcolor{blue}{...}} \\
    \textbf{\textcolor{blue}{Text:}} boring and obvious\\
    \textbf{\textcolor{blue}{Sentiment:}} Negative\\
    \textbf{\textcolor{blue}{Text:}} \textcolor{violet}{[Unlabeled Data]}\\
    \textbf{\textcolor{blue}{Sentiment:}} \textcolor{orange}{[Label]}
    }}
    };
    \end{tikzpicture}
    \vspace{-0.5em}
    \caption{An example of Prompt-Guided Unlabeled Data Annotation (PGDA) for SST2.}
    \label{fig:prompt_example_sst2}
\end{figure}

\subsection{Prompt-Guided Training Data Generation (PGDG)}
\label{sec:framework2}
The second approach is to utilize GPT-3 to autonomously generate labeled data for the specified task. This method involves the creation of prompts that guide GPT-3 to self-generate labeled data, which is subsequently used to train a local model to predict on human-labeled test data for the purpose of evaluation. For example, to generate training data with the relation "head of government", we can first "teach" GPT-3 to generate head-tail entity pairs that have the specified relation as illustrated in Figure~\ref{fig:prompt_example_fewrel1}. 
After we obtain the generated triplets (head-tail entity pairs with specified relation), as shown in Figure~\ref{fig:prompt_example_fewrel2}, we can then instruct GPT-3 to generate a sentence with the given entities and relation. Compared with tagging approach, a significant benefit of the generation-based approach is that it does not require a long list of label definitions specified in the prompt. For example, to generate NER data, it can first generate entities of each entity type (e.g. organization, person, etc.) and then generate a sentence with mixed entities.


\begin{figure}[t!]
    \centering
    \begin{tikzpicture}
    \node[draw, rounded corners] {
    \resizebox{0.8\linewidth}{!}{    
    \adjustbox{minipage=[r][13.5em][b]{0.46\textwidth},scale={0.7}}{
    \textbf{Generate 20 different Head Entity and Tail Entity with the given Relation.}\\
    \textbf{\textcolor{blue}{Relation:}} head of government\\
    \textbf{\textcolor{blue}{Relation Definition:}} head of the executive power of this town, city, municipality, state, country, or other governmental body\\
    \textbf{\textcolor{blue}{Relation:}} head of government\\
    \textbf{\textcolor{blue}{Head Entity:}} United States; \textbf{\textcolor{blue}{Tail Entity:}} Chester Alan Arthur\\
    \textbf{\textcolor{blue}{...}} \\
    \textbf{\textcolor{orange}{Head Entity:}} \textcolor{orange}{Entity1;} \textbf{\textcolor{orange}{Tail Entity:}} \textcolor{orange}{Entity2}
    }}
    };
    \end{tikzpicture}
    \vspace{-0.5em}
    \caption{An example of prompting GPT-3 to generate entities for the relation "head of government" for FewRel.}
    \label{fig:prompt_example_fewrel1}
\end{figure}

\begin{figure}[t!]
    \centering
    \begin{tikzpicture}
    \node[draw, rounded corners] {
    \resizebox{0.8\linewidth}{!}{    
    \adjustbox{minipage=[r][14.5em][b]{0.46\textwidth},scale={0.7}}{
    \textbf{Generate a sentence with the given entities and relation.}\\
    \textbf{\textcolor{blue}{Relation:}} head of government\\
    \textbf{\textcolor{blue}{Head Entity:}} United States; \textbf{\textcolor{blue}{Tail Entity:}} Chester Alan Arthur\\
    \textbf{\textcolor{blue}{Text:}} Chester Alan Arthur , 21st President of the United States , died of this disease , November 18 , 1886\\
    \textbf{\textcolor{blue}{...}} \\
    \textbf{\textcolor{blue}{Relation:}} head of government\\
    \textbf{\textcolor{violet}{Head Entity:}} \textcolor{violet}{Entity1;} \textbf{\textcolor{violet}{Tail Entity:}} \textcolor{violet}{Entity2}\\
    \textcolor{orange}{\textbf{Text:} [Generated Sentence]}
    }}
    };
    \end{tikzpicture}
    \vspace{-0.5em}
    \caption{An example of prompting GPT-3 to generate a sentence with the given entities and the relation "head of government" for FewRel.}
    \label{fig:prompt_example_fewrel2}
\end{figure}

\subsection{Dictionary-Assisted Training Data Generation (DADG)}
\label{sec:framework3}

The third method is designed to utilize a dictionary as an external source of knowledge to assist GPT-3 to generate labeled data for a specific domain.
In our experiments, we choose Wikidata\footnote{\url{https://www.wikidata.org}} as the dictionary. The data generated through this Wikidata-guided process is subsequently used to train a local model to predict human-labeled test data for the purpose of evaluating performance. For instance, to generate training data with the relation "head of government", we first query the head-tail entity pairs under the relation \emph{P6}, relation ID of ``head of government'', from Wikidata. Upon obtaining the entity pairs from Wikidata, GPT-3 can then be instructed to generate a sentence with the specified entity pairs and relation. An advantage of this approach is that it can leverage knowledge base in specific domains, particularly when the domains are not present in the pre-trained corpus, thus allowing for the incorporation of external knowledge into GPT-3 without the need for fine-tuning. 

%% file: 04_experiments.tex
\section{Experiments}
\label{sec:experiments}
\subsection{Experiment Settings}
In this study, we conduct extensive experiments on both sequence- and token-level NLP tasks\footnote{Please refer to Appendix \ref{sec:complex_tasks} for the discussion on more complex tasks like semantic parsing}. The sequence-level tasks include sentiment analysis (SA) and relation extraction (RE). The token-level tasks include named entity recognition (NER) and aspect sentiment triplet extraction (ASTE). 

More specifically, we use the SST2 dataset \cite{Socher2013RecursiveDM} for sentiment analysis, a well-known dataset comprising movie reviews. For relation extraction, we use FewRel \cite{han-etal-2018-fewrel}, a large-scale relation extraction dataset. For NER, we use the AI domain split from the CrossNER dataset \cite{Liu2020CrossNEREC}, which is the most difficult domain within the dataset and more closely mirrors real-world scenarios with its 14 entity types. For aspect sentiment triplet extraction, we use the laptop domain split released by \cite{xu-etal-2020-position}.

To simulate the production scenario, we assume that the user has access to the off-shelf GPT-3 API. In all our experiments, we use \emph{text-davinci-003}\footnote{Released on 28 Nov 2022. Please refer to \url{https://beta.openai.com/docs/models} for more details.}, the latest GPT-3 model. In addition, we assume that the user uses BERT$_\mathrm{BASE}$ for production and has access to a few data points and Wikidata for each task. For each task, the resulting data of each approach is post-processed and reformatted into the same format of human-labeled data before being used to fine-tune a BERT$_\mathrm{BASE}$ model. In order to accurately determine the cost and time required for human labeling, we conduct interviews and consultations with linguists and professional data annotators to obtain a precise estimation. 



\subsection{Sequence-Level Task}
\subsubsection{SST2}
SST2 dataset is used for sequence-level sentiment analysis experiments. We fine-tune BERT$_\mathrm{BASE}$ on the data created by the three approaches for 32 epochs with early stopping. After model fine-tuning, we evaluate the model on human-labeled test data to assess the quality of data created by each approach. We conduct experiments on zero-shot, 2-shot, and 10-shot settings. Here we discuss the results for 10-shot settings. Please refer to Appendix~\ref{sec:sst2_2shot} for the results of the other two settings.

\paragraph{Annotation Approaches}
In PGDA, we randomly sample 10-shot data of the train set of the SST2 dataset to construct a prompt template, as illustrated in Figure~\ref{fig:prompt_example_sst2}. The prompt is used to guide GPT-3 in generating sentiment labels for the unlabeled data. 
In PGDG, the same 10-shot data used in the PGDA is used to guide GPT-3 to generate sentences with specified sentiments. Please refer to Appendix~\ref{sec:sst2_pgdg} for the prompt example. 
In DADG, the ability of GPT-3 to perform Wikidata-guided few-shot generation is tested. We query entities in Wikidata from the movie domain. We then use the entities together with the same 10-shot data to prompt GPT-3 to generate sentences with a specified sentiment. Please refer to Appendix~\ref{sec:sst2_dadg} for the prompt example.

\paragraph{Results}
Table~\ref{tb:sst_result} presents the results of three different approaches. Overall, PGDA demonstrates the best performance among the three approaches. By labeling the same 3,000 data points, PGDA achieves an accuracy of 87.75, which is only 0.72 lower than that of human-labeled data. However, the cost and time consumed for PGDA are significantly lower than those for human labeling. By labeling 6,000 data, PGDA achieves a better performance than the human-labeled 3,000 data, while the cost is approximately 10\% of the cost of human labeling. PGDG performs much worse than PGDA and human-labeled data. However, it also demonstrates a distinct advantage in terms of cost and time efficiency when generating the same amount of data compared with alternative approaches. DADG approach, which involves generating data with in-domain entities, does not result in better performance. This is because entities are not typically key factors in the sentiment classification task, as most entities are neutral and do not provide additional information relevant to sentiment. Furthermore, since a large portion of the data in SST2 does not contain any entities, the sentences generated using DADG do not follow the same distribution as the test data in SST2, leading to poorer performance. For comparison purposes, the result of PGI is also presented. It is suggested that, for small-scale applications, it is practical to use GPT-3 to directly label unlabeled data.


\begin{table}[t!]
\centering
\scalebox{0.7}{
\begin{tabular}{lcccc}
\toprule

\textbf{Approach}              & \textbf{\begin{tabular}[c]{@{}c@{}}Num. of\\Samples\end{tabular}} & \textbf{\begin{tabular}[c]{@{}c@{}}Cost \\ (USD)\end{tabular}} & \textbf{\begin{tabular}[c]{@{}c@{}}Time \\ (Mins)\end{tabular}} & \textbf{Results} \\
\midrule

  \multirow{2}{*}{PGDA}          & 3000                                    & 11.31       & 14\dag          & 87.75          \\
                               & 6000                                    & 22.63       & 27\dag          & \textbf{89.29} \\
\midrule
\multirow{2}{*}{PGDG}          & 3000                                    & 0.91        & 4\dag           & 73.81          \\
                               & 6000                                    & 1.83        & 8\dag           & 76.55          \\
                                \midrule
\multirow{2}{*}{DADG}          & 3000                                    & 7.18        & 23\dag          & 68.04          \\
                               & 6000                                    & 14.37       & 46\dag          & 71.51          \\
\midrule 
\multirow{2}{*}{Human Labeled} & \cellcolor{palepink} 3000                                    & \cellcolor{palepink}221 - 300   & \cellcolor{palepink}1000        & \cellcolor{palepink}88.47          \\ 
                               & \cellcolor{palepink}67349 & \cellcolor{palepink}4800 - 6700 & \cellcolor{palepink}22740       & \cellcolor{palepink}93.52          \\

\midrule
PGI                               & \cellcolor{lavender}  1821                                                                               &\cellcolor{lavender} 7.33       & \cellcolor{lavender}12          & \cellcolor{lavender}95.77       \\ 
\bottomrule
\end{tabular}
}
 \vspace{-0.5em}
\caption{Costs, time spendings and results of SST2. \dag means multiprocessing (5 processes) is enabled. Time for manual labeling excludes the time spent on instruction preparation and training.}
\label{tb:sst_result}

\end{table}

\subsubsection{FewRel}
The FewRel dataset is used for RE experiments. The original FewRel dataset, proposed for meta-learning, is re-formulated to a supervised learning setting. The train data of FewRel, which comprises 64 distinct relations and 700 labeled instances for each relation, is divided into a new train/dev/test split (560/70/70). It is to simulate the real-world application of GPT-3 to annotate data for tasks with large label spaces. For FewRel experiments, we follow \cite{Devlin2019BERTPO} to fine-tune BERT$_\mathrm{BASE}$ on the data created by the three approaches for 3 epochs. Subsequently, the fine-tuned model is evaluated on the human-labeled test data to assess the quality of data produced by the proposed approaches. The number of samples annotated or generated by each approach is determined by assuring the costs of each approach are comparable.

\paragraph{Annotation Approaches}
The FewRel dataset poses significant challenges for the PGDA approach, primarily due to the complexity of instructing GPT-3 to comprehend the 64 relations. Due to the cost and maximum token length constraints of the GPT-3 API, we can only include 1-shot data for each relation within the prompt, which can make it difficult for GPT-3 to "understand" each relation. To address these challenges, we try 5 different prompts for PGDA, with the goal of exploring whether different prompts could be effective for tasks with large label space. Please refer to Appendix~\ref{sec:fewrel_pgda} for the prompt examples.
As mentioned in Section~\ref{sec:framework2}, in PGDG, we conduct the annotation for RE in two steps. The first step is to instruct GTP-3 to generate head-tail entity pairs for a specified relation and the second step is to generate sentences with the generated triplets. We generate 200 labeled data for each relation. As mentioned in Section~\ref{sec:framework3}, DADG for RE is also conducted in two steps. The first step is to query WikiData to obtain head-tail entity pairs for a specified relation and the second step is to generate sentences with the generated triplets. We generate 200 labeled data for each relation.

\paragraph{Results}

Table~\ref{tb:fewrel_result} presents the results of three different approaches. All five proposed prompts for PGDA perform badly on the FewRel task due to the task difficulty and large label space. In contrast, the generation-based approaches, namely PGDG and DADG, achieve much better performance with comparable costs. Even with access to only 1-shot data, PGDG and DADG yield F1 scores of around 44 and 40 points respectively in comparison to PGDA. With access to 5-shot data, the performances of PGDG and DADG are further improved with the increased diversity of the generated data. Under comparable costs, PGDG and DADG outperform the human-labeled data (704 data points) with 33-point and 23-point F1 scores respectively. It is worth noting that the PGDG approach consistently outperforms the DADG approach. Through analysis, it is determined that the head-tail entity pairs generated by PGDG possess greater diversity than those generated by DADG for specific relations such as religion and the language of the work. We do not perform PGI on FewRel data as the cost is obviously much higher.

\begin{table}[t!]
\centering
\scalebox{0.6}{
\begin{tabular}{lcccccc}
\toprule
\textbf{Approach} & \multicolumn{1}{c}{\textbf{\begin{tabular}[c]{@{}c@{}}Num. of\\Samples\end{tabular}}} & \multicolumn{1}{c}{\textbf{\begin{tabular}[c]{@{}c@{}}Cost \\ (USD)\end{tabular}}} & \multicolumn{1}{c}{\textbf{\begin{tabular}[c]{@{}c@{}}Time \\ (Mins)\end{tabular}}} & \multicolumn{1}{c}{\textbf{P}} & \multicolumn{1}{c}{\textbf{R}} & \multicolumn{1}{c}{\textbf{F1}} \\
\midrule
PGDA1 (1-shot)    & 384                                                                                         & 28.55               & 13\dag                   & 0.03               & 1.56            & 0.05        \\
PGDA2 (1-shot)    & 384                                                                                         & 25.40               & 10\dag                   & 0.14               & 1.7             & 0.18        \\
PGDA3 (1-shot)    & 384                                                                                         & 25.19               & 11\dag                   & 0.09               & 1.65            & 0.13        \\
PGDA4 (1-shot)    & 384                                                                                         & 25.57               & 10\dag                   & 0.02               & 1.56            & 0.05        \\
PGDA5 (1-shot)    & 384                                                                                         & 25.56               & 11\dag                   & 0.02               & 1.56            & 0.05        \\
PGDG (1-shot)     & 12800                                                                                       & 30.58               & 285\dag                  & 47.82              & 45.58           & \textbf{44.11}       \\
DADG (1-shot)     & 12800                                                                                       & 17.16               & 220\dag                  & 45.41              & 42.41           & 40.02       \\
\midrule
PGDG (5-shot)     & 12800                                                                                       & 99.35               & 340\dag                  & 70.59              & 67.99           & \textbf{67.71}       \\
DADG (5-shot)     & 12800                                                                                       & 88.91               & 265\dag                  & 59.76              & 60.85           & 57.98       \\
\midrule
\multirow{3}{*}{Human Labeled}      & \cellcolor{palepink} 704                                                                                         & \cellcolor{palepink}101 - 200              & \cellcolor{palepink}640                  & \cellcolor{palepink}41.92              & \cellcolor{palepink}41.45           &\cellcolor{palepink} 34.22       \\
    &\cellcolor{palepink} 12800                                                                                       &\cellcolor{palepink} 1828 - 3584             &\cellcolor{palepink} 11636                & \cellcolor{palepink}85.19              &\cellcolor{palepink} 85.07           &\cellcolor{palepink} 84.95       \\
     & \cellcolor{palepink} 35840                                                                                       & \cellcolor{palepink}6400 - 10,000            &\cellcolor{palepink} 32582                &\cellcolor{palepink} 87.55              &\cellcolor{palepink} 87.43           & \cellcolor{palepink} 87.34       \\
\midrule
 PGI               & \cellcolor{lavender} 4480                                                                                           & \cellcolor{lavender}33.30                   & \cellcolor{lavender} 160\dag                    &\cellcolor{lavender} 29.86                  & \cellcolor{lavender}29.82               &\cellcolor{lavender} 25.85         \\
\bottomrule
\end{tabular}
}
 \vspace{-0.5em}
\caption{Costs, time spendings, and results of  FewRel. Time for manual labeling excludes the time spent on instruction preparation and training. The number of samples annotated or generated by each approach is determined by assuring \textbf{comparable costs}. We use ChatGPT instead of GPT-3 to perform PGI on FewRel data as a proxy as the cost of using GPT-3 for PGI is obviously much higher. \dag means multiprocessing (5 processes) is enabled.} 
\label{tb:fewrel_result}
\end{table}

\subsection{Token-Level Task}
\subsubsection{CrossNER}
The AI domain split in CrossNER has 14 entity classes, namely product, field, task, researcher, university, programming language, algorithm, misc, metrics, organisation, conference, country, location, person. We fine-tune BERT$_\mathrm{BASE}$ on the CrossNER task with corresponding data for 100 epochs with early stopping. 

\paragraph{Annotation Approaches}
In PGDA, as shown in Appendix~\ref{sec:crossner_pgda}, for each entity type, we initiate GPT-3 to generate its definition and provide a selection of data (no more than 10-shot) with entities belonging to the specified entity type in the prompt to assist GPT-3 in recognizing entities belonging to the same class within the unlabeled data. It is observed that the same entity may be labeled as different entity types with different prompts. Therefore, we also include an additional prompt, as illustrated in Figure~\ref{fig:prompt_example5} in Appendix~\ref{sec:crossner_pgda}, to determine the final entity type for each identified entity. Both PGDG and DADG for CrossNER are conducted in two steps. The first step for PGDG is to prompt GPT-3 to generate entities for each entity type as shown in Appendix~\ref{sec:crossner_pgdg}. On the other hand, the first step for DADG is to query Wikidata to get the entities of each entity type. Notice that we use no more than 200 generated entities for each entity type in our experiments for both PGDG and DADG. The second step of both approaches is to use the generated entities to generate sentences within a specific domain using GPT-3 as shown in Figure~\ref{fig:prompt_example7} in Appendix~\ref{sec:crossner_pgda}. In the process of generating sentences for both PGDG and DADG, we randomly select a few entities from all the entities to generate each sentence.


 







\paragraph{Results}
Table~\ref{tb:crossNER_result} presents the results of the three approaches. We find the train data labeling method using PGDA has the worst performance yet the highest costs among the three proposed approaches. It should be noted that there are only 100 gold train data points in the AI domain split in the CrossNER dataset, and these same 100 data points are labeled using PGDA. However, the cost of labeling these 100 data points is higher than the cost of using the generation approaches to generate 3000 data points. It is observed that GPT-3 is effective at identifying entities in the text, but it may also identify entities that are not of the specified entity type, resulting in incorrect labeling. Additionally, GPT-3 may not accurately identify the boundaries of the entities. These two disadvantages make it impractical to use PGDA for labeling data for named entity recognition (NER) in a production setting, especially when the label space becomes bigger. The PGDG approach is able to achieve a result comparable to the 100 human-labeled gold train data at a lower cost. When utilizing Wikidata, the DADG approach is able to achieve a higher result than PGDG, likely due to its ability to leverage more unique entities and in-domain entities extracted from Wikidata. This shows that the ability to access in-domain entities is crucial for creating high-quality training data for NER.

\begin{table}[t!]
\centering
\scalebox{0.73}{
\begin{tabular}{lcccc}
\toprule
\textbf{Approach}              & \textbf{\begin{tabular}[c]{@{}c@{}}Num. of\\Samples\end{tabular}} & \textbf{\begin{tabular}[c]{@{}c@{}}Cost \\ (USD)\end{tabular}} & \textbf{\begin{tabular}[c]{@{}c@{}}Time \\ (Mins)\end{tabular}} & \textbf{Results} \\

\midrule
PGDA (10-shot)                    & 100                                                                                         & 15.39               & 21                   & 23.08            \\
\midrule
\multirow{2}{*}{PGDG (Zero-shot)} & 1500                                                                                        & 7.78                & 17\dag                    & 42.63            \\
                                  & 3000                                                                                        & 13.56               & 33\dag                    & 41.35            \\
\midrule
\multirow{2}{*}{DADG (Zero-shot)} & 1500                                                                                        & 6.77                & 20\dag                    & 46.90            \\
                                  & 3000                                                                                        & 13.61               & 40\dag                    & \textbf{47.22}            \\
\midrule
Human Labeled                     & \cellcolor{palepink} 100                                                                                         &\cellcolor{palepink} 17 - 42.85          & \cellcolor{palepink}65                   & \cellcolor{palepink}42.00            \\
\midrule
PGI                               & \cellcolor{lavender} 431                                                                                         &\cellcolor{lavender} 63.23               & \cellcolor{lavender}20\dag                    & \cellcolor{lavender}46.65           
 \\
\bottomrule
\end{tabular}
}
 \vspace{-0.5em}
\caption{Cost, time spending and results of CrossNER (AI Domain Split). Time for manual labeling excludes the time spent on instruction preparation and training. \dag means multiprocessing (5 processes) is enabled.}
\label{tb:crossNER_result}
\end{table}

\subsubsection{ASTE}
We follow \cite{xu-etal-2021-learning} to fine-tune BERT$_\mathrm{BASE}$ on the ASTE task using data created by each approach for 10 epochs and evaluate the fine-tuned models on human-labeled test data. We conduct our experiment under 10-shot settings.
\paragraph{Annotation Approaches}
In PGDA, we randomly sample 10-shot data from gold train data and use them to guide GPT-3 to tag the unlabeled data. Given the complexity of ASTE, which requires the identification of aspect, opinion, and sentiment triplets, we try 3 different prompts to assess the impact of different prompts on the overall performance of the tagging process. Please refer to Appendix~\ref{sec:aste_pgda} for more details. In PDGD, for comparison purposes, the same 10-shot data used for PGDA is used in the experiments for PGDG. We first instruct GPT-3 to generate aspect-opinion-sentiment triplets and then instruct GPT-3 to generate sentences with the generated triplets. We also try on 3 prompts under PGDG as specified in Appendix~\ref{sec:aste_pgdg}. In DADG, we query entities in laptop and computer hardware domains from WikiData and used them as aspects. We use the prompt that achieved the best performance for PGDG as the prompt to generate opinions and sentiments for the aspects. Then we use
the obtained triplets for sentence generation.


\paragraph{Results}
Table~\ref{tb:ASTE_result} presents the results of three different approaches. PGDA achieves the best performance compared with the other approaches. We also notice that performance varies with different prompts, which aligns with the previous research \cite{Luo2022BioGPTGP}. PGDG tends to generate data with explicit sentiment, as shown in Appendix~\ref{sec:gen_sample_aste}. Similar to SST2, as entities are not the key factors for ASTE and provide little help to this task, DADG is also outperformed by PGDA.

\begin{table}[t!]
\centering
\scalebox{0.63}{
\begin{tabular}{lcccccc}
\toprule

\textbf{Approach} & \multicolumn{1}{c}{\textbf{\begin{tabular}[c]{@{}c@{}}Num. of\\Samples\end{tabular}}} & \multicolumn{1}{c}{\textbf{\begin{tabular}[c]{@{}c@{}}Cost \\ (USD)\end{tabular}}} & \multicolumn{1}{c}{\textbf{\begin{tabular}[c]{@{}c@{}}Time \\ (Mins)\end{tabular}}} & \multicolumn{1}{c}{\textbf{P}} & \multicolumn{1}{c}{\textbf{R}} & \multicolumn{1}{c}{\textbf{F1}} \\

\midrule
PGDA1             & 906                                                                                         & 11.34               & 18                   & 57.93              & 44.38           & \textbf{50.26}       \\
PGDA2             & 906                                                                                         & 9.02                & 17                   & 50.78              & 24.13           & 32.71       \\
PGDA3             & 906                                                                                         & 12.84               & 19                   & 50.73              & 38.31           & 43.65       \\
\midrule
PGDG1             & 1000                                                                                        & 9.41                & 15\dag                   & 44.36              & 22.47           & 29.83       \\
PGDG2             & 1000                                                                                        & 7.68                & 14\dag                   & 54.93              & 14.36           & 22.77       \\
PGDG3             & 1000                                                                                        & 13.77               & 18\dag                   & 45.10              & 12.71           & 19.83       \\
\midrule
DADG              & 1000                                                                                        & 13.74               & 18\dag                   & 48.61              & 6.45            & 11.38       \\
\midrule
  \multirow{2}{*}{Human Labeled}   & \cellcolor{palepink}91                                                                                          & \cellcolor{palepink}13 - 20             & \cellcolor{palepink}180                  &\cellcolor{palepink} 45.14              &\cellcolor{palepink} 38.49           & \cellcolor{palepink}41.55       \\
    &\cellcolor{palepink} 906                                                                                         & \cellcolor{palepink}130 - 200           &\cellcolor{palepink} 1800                 & \cellcolor{palepink}63.07              & \cellcolor{palepink}55.99           &\cellcolor{palepink} 59.32       \\
\midrule
PGI               &  \cellcolor{lavender}328                                                                                         &\cellcolor{lavender} 3.92                & \cellcolor{lavender}9                    & \cellcolor{lavender}50.10              &\cellcolor{lavender} 48.43           & \cellcolor{lavender}49.25      
    \\
\bottomrule
\end{tabular}
}
 \vspace{-0.5em}
\caption{Costs, time spendings and results of ASTE (laptop domain split). Time for manual labeling excludes the time spent on instruction preparation and training. \dag means multiprocessing (5 processes) is enabled.}
\label{tb:ASTE_result}
\end{table}

%% file: 05_analysis.tex
\section{Further Analysis}
\label{sec:analysis}
\subsection{Impact of Label Space}

The results of our experiments indicate that the tagging-based approach (PGDA) is more appropriate for tasks with smaller label spaces and clearly defined labels. Examples of such tasks include sentence-level sentiment analysis and ASTE, which both have small label space (2-3 labels) that can be easily distinguished, e.g. positive, negative, neutral. In contrast, the generation-based approaches (PGDG and DADG) are better suited for tasks with larger label spaces or labels that possess a certain degree of ambiguity. Examples of such tasks include CrossNER and FewRel, which have 14 and 64\footnote{We refer to the train split of the FewRel used in our experiments. The original FewRel data has 100 labels in total.} labels respectively, and some of which may be difficult to identify or differentiate (e.g. Misc, etc.). Both the tagging-based and generation-based approaches have their own advantages and disadvantages. The tagging-based approach allows for direct access to in-domain unlabeled data, while the generation-based approaches may generate data that contains information that was "learned" during pre-training and may not align with the distribution of in-domain data. However, as the label space becomes larger, the tagging-based approach requires a lengthy prompt with examples to guide GPT-3, which can lead to catastrophic forgetting and increase annotation costs. On the other hand, the generation-based approaches can reformulate the task by first generating spans with labels (e.g. entities and triplets), and then generating a sentence with the labeled spans. These approaches reduce label errors and avoid the challenges of span boundary detection. In addition, generation-based approaches tend to be more cost-effective. as the prompts used can be significantly shorter when compared to those used in the tagging-based approach and multiple data can be generated with a single prompt at a time.

\subsection{Comparision with Human Annotators}
Through extensive experiments, we find that GPT-3 demonstrates promising ability to generate domain-specific data (e.g., entities in AI), structured data (e.g., triplets), as well as unstructured sequences at a fast speed. As discussed above, GPT-3 can even be used to generate data from scratch or to convert structured knowledge into natural sentences (Figure~\ref{fig:example}), eliminating the requirement of unlabeled data. While for human annotators, it usually takes longer time to train them for domain-specific data annotation, and their annotation speed is not comparable with machines in most cases. Moreover, it is often more challenging for humans to construct training data without unlabeled data, or when the size of label space is very large. Therefore, in terms of speed and domain-specific data annotation, and in the setting of labeled data generation, large language models (LLMs) exhibit encouraging potential. Machines are good at quickly labeling or generating a large amount of training data. However, if we limit the number of data samples for model training, the per-instance quality of the data annotated by humans is still higher in most cases.

\begin{figure}[t!]
    \centering
    \begin{tikzpicture}
    \node[draw, rounded corners] {
    \resizebox{0.7\linewidth}{!}{    
    \adjustbox{minipage=[r][6em][b]{0.46\textwidth},scale={0.5}}{
    \textbf{\textcolor{orange}{Generated Entities:}} Chiang Mai International Airport; Chiang Mai, Thailand; \\
    \textbf{\textcolor{purple}{Generated Sentence:}} Chiang Mai International Airport is the main gateway for air travels to and from Chiang Mai, Thailand.
    }}
    };
    \end{tikzpicture}
    \vspace{-0.5em}
    \caption{An example to demonstrate the generation ability of GPT-3.}
    \label{fig:example}
\end{figure}

\subsection{Impact of Number of Shots}
\begin{figure}[t!]
    \centering
    \includegraphics[scale=0.5]{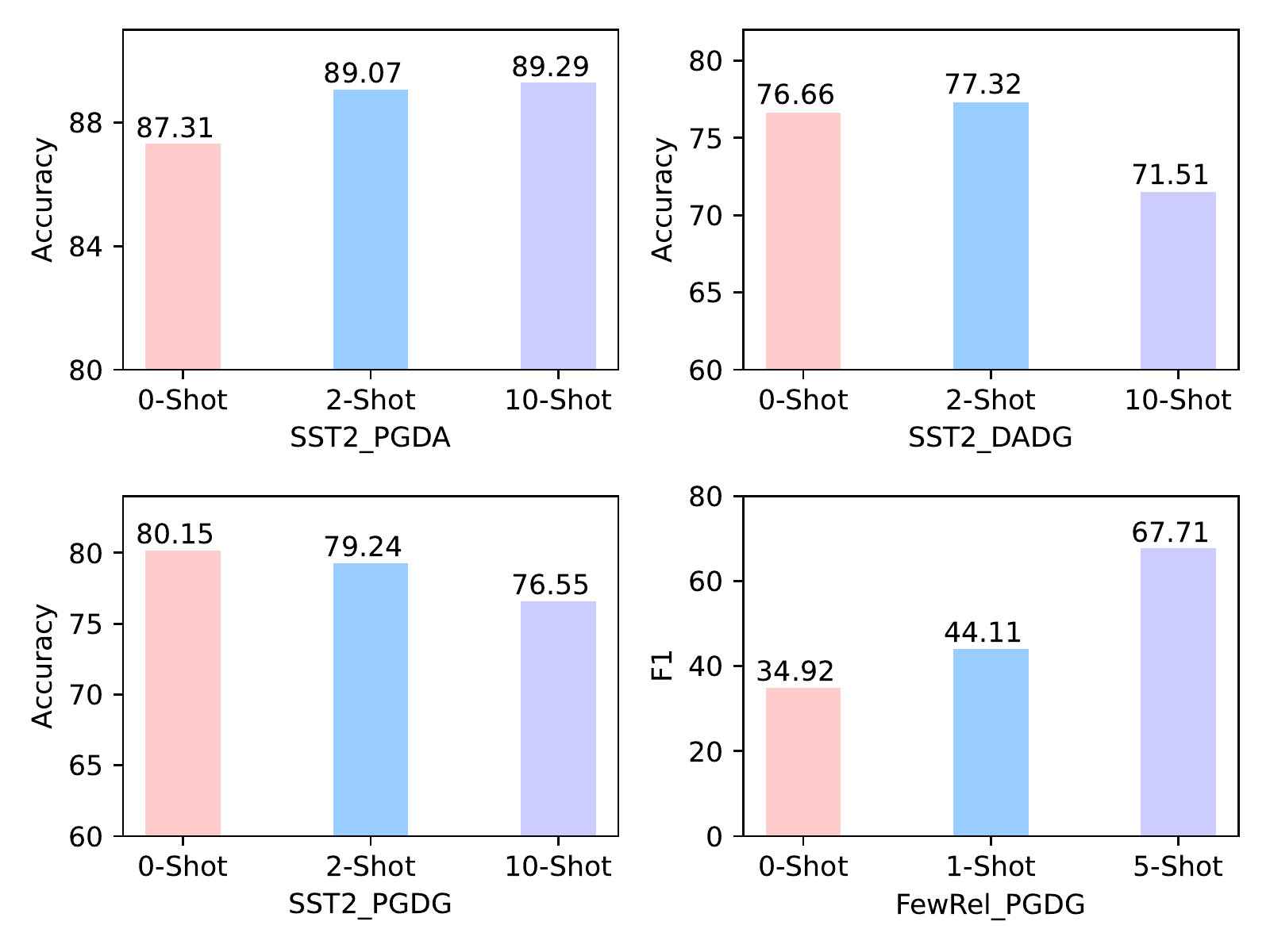}
     \vspace{-1.5em}
    \caption{Experiments on the impact of number of shots. We reported the results of 6,000 data on SST2 and 12,800 data (200 data per class) on FewRel.}
    \label{fig:num_shot} 
\end{figure}

We conduct experiments on the following two datasets, SST2 and FewRel to explore the impact of the number of shots. We find that increasing the number of shots does not necessarily lead to better annotation results for all approaches. As shown in Figure~\ref{fig:num_shot}, for SST2, tagging approach (PGDA) can benefit from more examples in the context, which enhances GPT-3's ability to tag unlabeled data. However, for the PGDG and DADG approaches, GPT-3 tends to generate data similar to the given examples. As shown in Figure~\ref{fig:case_study}, for SST2, the data is usually not a complete sentence and tend to be short and carry less information. Thus, with more data examples, GPT-3 will ``learn'' to generate similar data with less information and lead to poorer data quality. However, for FewRel, the data is a complete sentence and carry lots of information and the relations between the head entity and tail entity tend to be more implicit. Thus, with 5-shot data in the context, GPT-3 can generate data that also contain more implicit relations than only with 1-shot or zero-shot in the context\footnote{Please refer to Appendix~\ref{sec:gen_sample_for_sst2_fewrel} for the examples of generated data with different number of shots for SST2 and FewRel.}.

\subsection{Preliminary Comparison between GPT-3 and ChatGPT}

\begin{table}[t!]
\centering
\scalebox{0.73}{
\begin{tabular}{lccc}
\toprule
\textbf{Model}   & \textbf{Numb. of Sampes} & \textbf{Cost}  & \textbf{Results}              \\
\midrule
 GPT-3   & 3000             & 11.31 & \textbf{87.75}                \\
 ChatGPT & 3000             & \textbf{1.50}  & 87.31                \\
\bottomrule
\end{tabular}
}
\caption{Preliminary Comparison between GPT-3 and ChatGPT on SST2.}
\label{tb:comparison_sst2}
\end{table}

Based on the findings presented in Table~\ref{tb:comparison_sst2}, our analysis reveals that ChatGPT exhibits a performance level that is on par with GPT-3 when it comes to the SST2 task. Notably, the results obtained from our observations demonstrate comparable outcomes between ChatGPT and GPT-3 in terms of task performance. Moreover, from a cost-efficiency standpoint, ChatGPT emerges as a more economically viable alternative when compared to GPT-3, which may make it a preferable choice. A study conducted by \citet{gilardi2023chatgpt} further illustrates the superior performance of ChatGPT compared to crowd-workers for various annotation tasks. By employing a dataset consisting of 2,382 tweets, the research demonstrates that ChatGPT surpasses the capabilities of crowd-workers across multiple annotation tasks, including relevance assessment, stance analysis, topic identification, and frame detection. These findings suggest that large language models may outperform human annotators when it comes to these specific tasks, highlighting their potential as a highly effective and reliable tool for annotation purposes.


\subsection{Case Study on Multilingual Data Annotation}
As shown in Appendix~\ref{sec:multilingual}, we meticulously examined the annotation capabilities of state-of-the-art language models, namely GPT-3, ChatGPT, and GPT-4, within the context of multilingual training data. Our observations revealed that these models possess the remarkable ability to annotate such data effectively, even when presented with minimal or no prior exposure to the target languages. By employing a zero shot or few shot setting, where the models were not explicitly fine-tuned on the specific languages in question, we witnessed their capacity to accurately annotate and comprehend diverse linguistic inputs from a multitude of languages. This notable achievement underscores the potential of these language models to transcend language barriers and facilitate efficient multilingual data processing, making them invaluable tools for a wide range of language-related tasks and applications.

\begin{figure}[t!]
    \centering
    \begin{tikzpicture}
    \node[draw, rounded corners] {
    \resizebox{0.7\linewidth}{!}{    
    \adjustbox{minipage=[r][6em][b]{0.46\textwidth},scale={0.5}}{
    \textbf{\textcolor{orange}{SST Example:}} a smile on your face (positive)\\
    \textbf{\textcolor{purple}{FewRel Example:}} Winscombe is a lightly populated locality in the southern part of the Canterbury region of New Zealand 's South Island . (Relation: located on terrain feature)
    }}
    };
    \end{tikzpicture}
    \vspace{-0.5em}
    \caption{Examples to show the differences between the data distributions of SST2 and FewRel data.}
    \label{fig:case_study}
\end{figure}

%% file: 06_conclusion.tex
\section{Conclusions}
\label{sec:conclusion}
 In this work, we investigate the effectiveness of GPT-3 as a data annotator for various natural language processing (NLP) tasks using three main approaches. Our experimental results show that GPT-3 has the potential to annotate data for different tasks at a relatively lower cost, especially for individuals or organizations with limited budgets. With the limited budget, performance of model trained on the GPT-3 annotated data is often comparable to or even better than that trained on human-annotated data. However, it should be noted that the quality of data annotated by GPT-3 still has room for improvement when compared to human-annotated data. We hope the findings in this work can shed the light on automatic data annotation using large language models and provide some insights so that more methods can be proposed to enhance the quality of data created by these models. With everyone being able to create data for their model training, we can pave the way for the democratization of AI.

\section*{Acknowledgements}
\label{sec:acknowledgements}
This research is supported, in part, by Alibaba Group through Alibaba Innovative Research (AIR) Program and Alibaba-NTU Singapore Joint Research Institute (JRI), Nanyang Technological University, Singapore. We would like to thank Aljunied Mahani for her contributions on her valuable advice in this project from a linguist perspective.

I (Bosheng) would like to express my heartfelt gratitude and deepest appreciation to the memory of my beloved father, whose untimely departure last year has left an indelible void in my life. His unwavering support, encouragement, and wisdom were instrumental in shaping my character and nurturing my academic pursuits. I am forever indebted to his enduring love and belief in my abilities. Though he is no longer physically present, his spirit continues to guide and inspire me as I embark on this research journey. This work is dedicated to his cherished memory, a constant reminder of his profound impact on my life.


\section{Limitations}
Our work is subject to certain limitations, one of which pertains to financial constraints that hindered the ability to conduct large-scale experimentation with the data annotation methods proposed. As a result, the findings of this study may not be fully representative of larger datasets or populations. Additionally, the utilization of GPT-3 as a model presents challenges in terms of interpretability, as it operates as a "black box" system. To further investigate this subject, it would be beneficial to conduct larger-scale experiments and to compare the performances of GPT-3, ChatGPT\footnote{\href{https://chat.openai.com/chat} https://chat.openai.com/chat}, and GPT-4 \cite{openai2023gpt} and the open-sourced LLMs like LLaMA \cite{touvron2023llama}.

\section*{Ethics Consideration}
One of the significant issues associated with GPT-3 is the potential for it to reinforce existing biases present in the data sets it annotated. This is due to GPT-3 being pre-trained on a vast amount of unlabelled data, which may include bias and stereotypes \cite{Li2022DoesGD}. To address this concern, it is crucial to guarantee that the data used to train GPT-3 is diverse and representative of various viewpoints and experiences. Furthermore, consistent monitoring and evaluation of the output generated by GPT-3 should be implemented to identify and rectify any possible biases.

%% file: 07_appendix.tex
\section{Appendix}

\subsection{PGDA for SST2} 
\label{sec:sst2_pgda}
\begin{figure}[ht!]
    \centering
    \begin{tikzpicture}
    \node[draw, rounded corners] {
    \resizebox{0.8\linewidth}{!}{    
    \adjustbox{minipage=[r][11em][b]{0.46\textwidth},scale={0.7}}{
    \textbf{Choose the sentiment of the given text from Positive and Negative.}\\
    \textbf{\textcolor{blue}{Text:}} a feast for the eyes\\
    \textbf{\textcolor{blue}{Sentiment:}} Positive\\
     \textbf{\textcolor{blue}{...}} \\
    \textbf{\textcolor{blue}{Text:}} boring and obvious\\
    \textbf{\textcolor{blue}{Sentiment:}} Negative\\
    \textbf{\textcolor{blue}{Text:}} \textcolor{violet}{[Unlabeled Data]}\\
    \textbf{\textcolor{blue}{Sentiment:}} \textcolor{orange}{[Label]}
    }}
    };
    \end{tikzpicture}
    \caption{An example of prompt-guided unlabeled data annotation for SST2.}
    \label{fig:prompt_example_sst2_1}
\end{figure}

\subsection{PGDG for SST2} 
\label{sec:sst2_pgdg}
\begin{figure}[ht!]
    \centering
    \begin{tikzpicture}
    \node[draw, rounded corners] {
    \resizebox{0.8\linewidth}{!}{    
    \adjustbox{minipage=[r][9em][b]{0.46\textwidth},scale={0.7}}{
    \textbf{Write 20 different movie reviews with positive sentiments with no more than 20 words.}\\
     \textbf{\textcolor{blue}{Sentiment:}} Positive\\
    \textbf{\textcolor{blue}{Text:}} a feast for the eyes\\
     \textbf{\textcolor{blue}{...}} \\
    \textbf{\textcolor{blue}{Sentiment:}} Positive\\
    \textbf{\textcolor{blue}{Text:}}
    }}
    };
    \end{tikzpicture}
    \caption{An example of prompt-guided data generation for SST2.}
    \label{fig:fig:prompt_example_sst2_2}
\end{figure}

\subsection{DADG for SST2} 
\label{sec:sst2_dadg}
\begin{figure}[ht!]
    \centering
    \begin{tikzpicture}
    \node[draw, rounded corners] {
    \resizebox{0.8\linewidth}{!}{    
    \adjustbox{minipage=[r][10em][b]{0.46\textwidth},scale={0.7}}{
    \textbf{\textcolor{blue}{Sentiment:}} Positive\\
    \textbf{\textcolor{blue}{Text:}} a feast for the eyes\\
     \textbf{\textcolor{blue}{...}} \\
    \textbf{Write a movie review with the given entity with positive sentiment.}\\
     \textbf{\textcolor{blue}{Entity:}} [Entity1]\\
    \textbf{\textcolor{blue}{Sentiment:}} Positive\\
    \textbf{\textcolor{blue}{Text:}}
    }}
    };
    \end{tikzpicture}
    \caption{An example of dictionary-assisted training data
generation for SST2.}
    \label{fig:prompt_example_sst2_3}
\end{figure}

\newpage
\subsection{PGDA for CrossNER}
\label{sec:crossner_pgda}


\begin{figure}[ht!]
    \centering
    \begin{tikzpicture}
    \node[draw, rounded corners] {
    \resizebox{0.8\linewidth}{!}{    
    \adjustbox{minipage=[r][18em][b]{0.46\textwidth},scale={0.7}}{
     \textbf{\textcolor{blue}{Researcher:}} A researcher in AI domain is an individual who conducts research and experiments related to Artificial Intelligence and its related fields, such as ... \\\\
    \textbf{\textcolor{blue}{Text:}} Advocates of procedural representations were mainly centered at MIT , under the leadership of Marvin Minsky and Seymour Papert . \\
    \textbf{\textcolor{blue}{ Researcher entity:}} Marvin Minsky; Seymour Papert;\\
    \textbf{\textcolor{blue}{...}}\\
    \textbf{\textcolor{blue}{Text:}} [Unlabeled Data]\\
    \textbf{\textcolor{blue}{ Researcher entity:}}
    }}
    };
    \end{tikzpicture}
    \caption{An example of prompt-guided unlabeled data
annotation for CrossNER.}
    \label{fig:prompt_example_crossNER_pgda}
\end{figure}

\begin{figure}[ht!]
    \centering
    \begin{tikzpicture}
    \node[draw, rounded corners] {
    \resizebox{0.8\linewidth}{!}{    
    \adjustbox{minipage=[r][22em][b]{0.46\textwidth},scale={0.7}}{
    \textbf{Choose the right entity type from the candidate list for the given entity in the text context.}\\
    \textbf{\textcolor{blue}{Text:}} Advocates of procedural representations were mainly centered at MIT, under the leadership of Marvin Minsky and Seymour Papert .\\
    \textbf{\textcolor{blue}{Entity:}} Marvin Minsky\\
    \textbf{\textcolor{blue}{Candidate List:}} product, task, researcher, university, organisation, person\\
    \textbf{\textcolor{blue}{Entity Type:}} researcher\\
    \textbf{\textcolor{blue}{...}}\\
    \textbf{\textcolor{blue}{Text:}} [Unlabeled Data]\\
    \textbf{\textcolor{blue}{Entity:}} [Entity]\\
    \textbf{\textcolor{blue}{Candidate List:}} [Entity\_Type1, Entity\_Type2, Entity\_Type3, ...]\\
    \textbf{\textcolor{blue}{Entity Type:}}\\
    }}
    };
    \end{tikzpicture}
    \caption{An example of prompt to determine the entity type of an entity in CrossNER.}
    \label{fig:prompt_example5}
\end{figure}

\newpage
\subsection{PGDG and DADG for CrossNER}
\label{sec:crossner_pgdg}

\begin{figure}[ht!]
    \centering
    \begin{tikzpicture}
    \node[draw, rounded corners] {
    \resizebox{0.8\linewidth}{!}{    
    \adjustbox{minipage=[r][16em][b]{0.46\textwidth},scale={0.7}}{
    \textbf{\textcolor{blue}{Researcher:}} A researcher in AI domain is an individual who conducts research and experiments related to Artificial Intelligence and its related fields, such as Machine Learning ...\\
    \textbf{\textcolor{blue}{Researcher:}} David Silver, Fei-Fei Li, Claude Shannon, Marvin Minsky, Ruslan Salakhutdinov\\
    Generate 15 different researchers in the AI domain.\\
    \textbf{\textcolor{blue}{Researcher:}}\\
    1. David Silver\\
    2. ... \\
    }}
    };
    \end{tikzpicture}
    \caption{An example of prompting GPT-3 to generate entities for the type `Researcher' for PGDG.}
    \label{fig:prompt_example6}
\end{figure}

\begin{figure}[ht!]
    \centering
    \begin{tikzpicture}
    \node[draw, rounded corners] {
    \resizebox{0.9\linewidth}{!}{    
    \adjustbox{minipage=[r][8em][b]{0.46\textwidth},scale={0.7}}{
    \textbf{Generate text with all the given entities in the AI domain.}\\
    \textbf{\textcolor{blue}{Entities:}} Entity1$\_$Type: Entity1; Entity2$\_$Type: Entity2; ...\\
    \textbf{\textcolor{blue}{Text:}}\\
    }}
    };
    \end{tikzpicture}
    \caption{An example of prompting GPT-3 to generate a sentence with given entities for both PGDG and DADG.}
    \label{fig:prompt_example7}
\end{figure}

\newpage
\subsection{Generated Samples for ASTE by GPT-3}
\label{sec:gen_sample_aste}
\begin{figure}[h!]
    \centering
    \begin{tikzpicture}
    \node[draw, rounded corners] {
    \resizebox{0.9\linewidth}{!}{    
    \adjustbox{minipage=[r][10em][b]{0.46\textwidth},scale={0.5}}{
    \textbf{\textcolor{orange}{Gold train data:}} The biggest problem is that the box had no instructions in it .\\
    \textbf{\textcolor{purple}{Data generated by PGDG:}} The port layout is good and the processor is good for the price .\\
    \textbf{\textcolor{red}{Data generated by DADG:}} The Edge device is quite lightweight , the PC speaker is mediocre, but great for a Toshiba T3100 and good for other peripherals. 
    }}
    };
    \end{tikzpicture}
    \caption{Examples to compare the gold train data and the sentences generated by GPT-3. GPT-3 tends to generate data with more explicit sentiment expressions compared with gold train data.}
    \label{fig:case_study1}
\end{figure}

\subsection{Generated Samples for SST2 and FewRel for Different Number of Shots}
\label{sec:gen_sample_for_sst2_fewrel}
\begin{figure}[h!]
    \centering
    \begin{tikzpicture}
    \node[draw, rounded corners] {
    \resizebox{0.9\linewidth}{!}{    
    \adjustbox{minipage=[r][8em][b]{0.46\textwidth},scale={0.5}}{
    \textbf{\textcolor{orange}{Zero-shot:}} Fantastic! Great performances, an incredible soundtrack, and a captivating plot.\\
    \textbf{\textcolor{purple}{1-shot:}} A heartfelt and sincere film that will leave you feeling uplifted\\
     \textbf{\textcolor{red}{5-shot:}} a real crowd-pleaser\\
    }}
    };
    \end{tikzpicture}
    \vspace{-0.5em}
    \caption{Examples to show the sentences generated by GPT-3 under Zero-shot, 1-shot, and 5-shot settings for SST2 with PDPG.}
    \label{fig:case_study1}
\end{figure}

\begin{figure}[h!]
    \centering
    \begin{tikzpicture}
    \node[draw, rounded corners] {
    \resizebox{0.9\linewidth}{!}{    
    \adjustbox{minipage=[r][12em][b]{0.46\textwidth},scale={0.5}}{
    \textbf{\textcolor{orange}{Zero-shot:}} The Dallas Airport is a transport hub that serves the city of Dallas.\\
    \textbf{\textcolor{purple}{1-shot:}} Narita Airport ( NRT ) serves as the main transport hub for flights to and from Narita.\\
     \textbf{\textcolor{red}{5-shot:}} It serves as Manila's main international gateway , being located at the heart of Manila International Airport Complex at Ninoy Aquino International Airport in Manila , Philippines.\\
    }}
    };
    \end{tikzpicture}
    \vspace{-0.5em}
    \caption{Examples to show the sentences generated by GPT-3 under Zero-shot, 1-shot, and 5-shot settings for FewRel with PDPG.}
    \label{fig:case_study2}
\end{figure}

\newpage

\newpage

\subsection{PGDA for ASTE}
\label{sec:aste_pgda}

\begin{figure}[ht!]
    \centering
    \begin{tikzpicture}
    \node[draw, rounded corners] {
    \resizebox{0.8\linewidth}{!}{    
    \adjustbox{minipage=[r][15em][b]{0.46\textwidth},scale={0.7}}{
    \textbf{Identify the target, opinion, and sentiment triplets in the given text.}\\
    \textbf{\textcolor{blue}{Text:}} The biggest problem is that the box had no instructions in it . \\
    \textbf{\textcolor{blue}{Target0:}} instructions; Opinion0: problem; Sentiment0: negative \\
    \textbf{\textcolor{blue}{Target1:}} instructions; Opinion1: no; Sentiment1: negative \\
    \textbf{\textcolor{blue}{...}} \\
    \textbf{\textcolor{blue}{Text:}} \textcolor{violet}{[Unlabeled Data]}\\
    \textbf{\textcolor{blue}{Target0:}} \textcolor{orange}{[Label]} ...\\
    }}
    };
    \end{tikzpicture}
    \caption{Prompt for PGDA1 for ASTE.}
    \label{fig:prompt_aste1}
\end{figure}

\begin{figure}[ht!]
    \centering
    \begin{tikzpicture}
    \node[draw, rounded corners] {
    \resizebox{0.8\linewidth}{!}{    
    \adjustbox{minipage=[r][14em][b]{0.46\textwidth},scale={0.7}}{
    \textbf{Identify the target, opinion, and sentiment triplets in the given text.}\\
    \textbf{\textcolor{blue}{Text:}} The biggest problem is that the box had no instructions in it . \\
    \textbf{\textcolor{blue}{Target:}}instructions; instructions;\\
    \textbf{\textcolor{blue}{Opinion:}} problem; no;\\
    \textbf{\textcolor{blue}{Sentiment:}} negative; negative;\\
    \textbf{\textcolor{blue}{...}} \\
    \textbf{\textcolor{blue}{Text:}} \textcolor{violet}{[Unlabeled Data]}\\
    \textbf{\textcolor{blue}{Target:}} \textcolor{orange}{[Label]}, ...\\
    }}
    };
    \end{tikzpicture}
    \caption{Prompt for PGDA2 for ASTE.}
    \label{fig:prompt_aste2}
\end{figure}

\begin{figure}[ht!]
    \centering
    \begin{tikzpicture}
    \node[draw, rounded corners] {
    \resizebox{0.8\linewidth}{!}{    
    \adjustbox{minipage=[r][15em][b]{0.46\textwidth},scale={0.7}}{
    \textbf{Identify the target, opinion, and sentiment triplets in the given text.}\\
    \textbf{\textcolor{blue}{Text:}} The biggest problem is that the box had no instructions in it .\\
    \textbf{\textcolor{blue}{Target0:}} is instructions. Its opinion span is problem. Its sentiment is negative.\\
    \textbf{\textcolor{blue}{Target1:}} is instructions. Its opinion span is no. Its sentiment is negative.\\
    \textbf{\textcolor{blue}{...}} \\
    \textbf{\textcolor{blue}{Text:}} \textcolor{violet}{[Unlabeled Data]}\\
    \textbf{\textcolor{blue}{Target0:}} is \textcolor{orange}{[Label]} ...\\
    }}
    };
    \end{tikzpicture}
    \caption{Prompt for PGDA3 for ASTE.}
    \label{fig:prompt_aste3}
\end{figure}

\newpage
\subsection{PGDG and DADG for ASTE}
\label{sec:aste_pgdg}

\begin{figure}[ht!]
    \centering
    \begin{tikzpicture}
    \node[draw, rounded corners] {
    \resizebox{0.7\linewidth}{!}{    
    \adjustbox{minipage=[r][10em][b]{0.46\textwidth},scale={0.7}}{
    \textbf{Generate 20 different sentiment, target and opinion triplets.}\\
    \textbf{\textbf{\textcolor{blue}{1. Target0:}}} instructions; Opinion0: problem; Sentiment0: negative; Target1: instructions; Opinion1: no; Sentiment1: negative;\\
    \textbf{\textcolor{blue}{...}} \\
    \textbf{\textcolor{blue}{Target0:}} \textcolor{orange}{[Target0]} ...\\
    }}
    };
    \end{tikzpicture}
    \caption{Prompt for PGDG1 for ASTE.}
    \label{fig:prompt_aste4}
\end{figure}

\begin{figure}[ht!]
    \centering
    \begin{tikzpicture}
    \node[draw, rounded corners] {
    \resizebox{0.7\linewidth}{!}{    
    \adjustbox{minipage=[r][9em][b]{0.46\textwidth},scale={0.7}}{
    \textbf{Generate 20 different sentiment, target and opinion triplets.}\\
    \textbf{\textcolor{blue}{1. Target:}}  instructions; instructions; Opinion: problem; no; Sentiment: negative; negative;\\
    \textbf{\textcolor{blue}{...}} \\
    \textbf{\textcolor{blue}{11. Target:}} \textcolor{orange}{[Target0]}; ...\\
    }}
    };
    \end{tikzpicture}
    \caption{Prompt for PGDG2 for ASTE.}
    \label{fig:prompt_aste5}
\end{figure}

\begin{figure}[ht!]
    \centering
    \begin{tikzpicture}
    \node[draw, rounded corners] {
    \resizebox{0.7\linewidth}{!}{    
    \adjustbox{minipage=[r][10em][b]{0.46\textwidth},scale={0.7}}{
    \textbf{Generate 20 different targets and opinions in positive sentiment.}
    \textbf{\textcolor{blue}{Sentiment:}} positive; Target: features; Opinion: nice; \\
    \textbf{\textcolor{blue}{Sentiment:}} positive; Target: priced; Opinion: reasonable;\\
    \textbf{\textcolor{blue}{...}} \\
    \textbf{\textcolor{blue}{Sentiment:}} positive; Target:\textcolor{orange}{[Target0]} ...\\
    }}
    };
    \end{tikzpicture}
    \caption{Prompt for PGDG3 for ASTE.}
    \label{fig:prompt_aste6}
\end{figure}
\begin{figure}[ht!]
    \centering
    \begin{tikzpicture}
    \node[draw, rounded corners] {
    \resizebox{0.7\linewidth}{!}{    
    \adjustbox{minipage=[r][18em][b]{0.46\textwidth},scale={0.7}}{
    \textbf{Generate a sentence with the given target, opinion and sentiment triplets in the laptop domain.}\\
    \textbf{\textcolor{blue}{Target0:}} instructions; Opinion0: problem; Sentiment0: negative; Target1: instructions; Opinion1: no; Sentiment1: negative;\\
    \textbf{\textcolor{blue}{Text:}} The biggest problem is that the box had no instructions in it .\\
    \textbf{\textcolor{blue}{...}} \\
    \textbf{\textcolor{blue}{Target0:}} [Target0]; Opinion0: [Opinion0]; Sentiment0: [Sentiment0];\\
    \textbf{\textcolor{blue}{...}} \\
    \textbf{\textcolor{blue}{Text:}} \textcolor{orange}{[Generated Sentence]}\\
    }}
    };
    \end{tikzpicture}
    \caption{An example of Prompting GPT-3 to generate a sentence with given triplets for ASTE using PGDG and DADG.}
    \label{fig:prompt_aste7}
\end{figure}

\newpage
\subsection{PGDA for FewRel}
\label{sec:fewrel_pgda}
\begin{figure}[ht!]
    \centering
    \begin{tikzpicture}
    \node[draw, rounded corners] {
    \resizebox{0.9\linewidth}{!}{    
    \adjustbox{minipage=[r][52em][b]{0.46\textwidth},scale={0.7}}{
    \textbf{Identify the relation between the head entity and the tail entity in the given sentence.}\\
    \textbf{\textcolor{blue}{Relation:}} place served by transport hub; mountain range; religion; participating team; contains administrative territorial entity; head of government; country of citizenship; original network; heritage designation; performer; participant of; position held; has part; location of formation; located on terrain feature; architect; country of origin; publisher; director; father; developer; military branch; mouth of the watercourse; nominated for; movement; successful candidate; followed by; manufacturer; instance of; after a work by; member of political party; licensed to broadcast to; headquarters location; sibling; instrument; country; occupation; residence; work location; subsidiary; participant; operator; characters; occupant; genre; operating system; owned by; platform; tributary; winner; said to be the same as; composer; league; record label; distributor; screenwriter; sports season of league or competition; taxon rank; location; field of work; language of work or name; applies to jurisdiction; notable work; located in the administrative territorial entity;\\
    \\
    \textbf{\textcolor{blue}{Sentence:}} Merpati flight 106 departed Jakarta ( CGK ) on a domestic flight to Tanjung Pandan ( TJQ ) .
    \textbf{\textcolor{blue}{Head Entity:}} TJQ; Tail Entity: Tanjung Pandan\\
    \textbf{\textcolor{blue}{Relation:}} place served by transport hub\\
    \textbf{\textcolor{blue}{Sentence:}} It is approximately 8 km away from Mount Korbu , the tallest mountain of the Titiwangsa Mountains .\\
    \textbf{\textcolor{blue}{Head Entity:}} Mount Korbu; Tail Entity: Titiwangsa Mountains\\
    \textbf{\textcolor{blue}{...}} \\
    \textbf{\textcolor{blue}{Sentence1:}} [unlabeled data]\\
    \textbf{\textcolor{blue}{Head Entity1:}} [head entity]; Tail Entity1:[tail entity]\\
    \textbf{\textcolor{blue}{Relation:}}  \textcolor{orange}{[label]}\\
    }}
    };
    \end{tikzpicture}
    \caption{Prompt for PGDA1 used for FewRel Experiemtns.}
    \label{fig:prompt1forfewrel}
\end{figure}

\newpage
\begin{figure}[ht!]
    \centering
    \begin{tikzpicture}
    \node[draw, rounded corners] {
    \resizebox{0.9\linewidth}{!}{    
    \adjustbox{minipage=[r][56em][b]{0.46\textwidth},scale={0.7}}{
    \textbf{Identify the relation between the head entity and the tail entity in the given sentence.}\\
    \textbf{\textcolor{blue}{Relation:}} place served by transport hub; mountain range; religion; participating team; contains administrative territorial entity; head of government; country of citizenship; original network; heritage designation; performer; participant of; position held; has part; location of formation; located on terrain feature; architect; country of origin; publisher; director; father; developer; military branch; mouth of the watercourse; nominated for; movement; successful candidate; followed by; manufacturer; instance of; after a work by; member of political party; licensed to broadcast to; headquarters location; sibling; instrument; country; occupation; residence; work location; subsidiary; participant; operator; characters; occupant; genre; operating system; owned by; platform; tributary; winner; said to be the same as; composer; league; record label; distributor; screenwriter; sports season of league or competition; taxon rank; location; field of work; language of work or name; applies to jurisdiction; notable work; located in the administrative territorial entity;\\
    \\
    \textbf{\textcolor{blue}{Sentence:}} Merpati flight 106 departed Jakarta ( CGK ) on a domestic flight to Tanjung Pandan ( TJQ ) .
    \textbf{\textcolor{blue}{the relation between}} TJQ and Tanjung Pandan is place served by transport hub\\
    \textbf{\textcolor{blue}{Sentence:}} It is approximately 8 km away from Mount Korbu , the tallest mountain of the Titiwangsa Mountains .\\
    \textbf{\textcolor{blue}{the relation between}} Mount Korbu and Titiwangsa Mountains is mountain range\\
    \textbf{\textcolor{blue}{Sentence:}} In 1689 , Konstanty was one of the judges who sentenced Kazimierz Łyszczyński to death for atheism .\\
    \textbf{\textcolor{blue}{the relation between}} Kazimierz Łyszczyński and atheism is religion\\
    ...\\
    \textbf{\textcolor{blue}{Sentence1:}} \textcolor{violet}{[unlabeled data]}\\
    \textbf{\textcolor{blue}{the relation between}} [head entity] and [tail entity] is \textcolor{orange}{[label]}\\
    }}
    };
    \end{tikzpicture}
    \caption{Prompt for PGDA2 used for FewRel Experiemtns.}
    \label{fig:prompt2forfewrel}
\end{figure}

\newpage
\begin{figure}[ht!]
    \centering
    \begin{tikzpicture}
    \node[draw, rounded corners] {
    \resizebox{0.9\linewidth}{!}{    
    \adjustbox{minipage=[r][48em][b]{0.46\textwidth},scale={0.7}}{
    \textbf{Identify the relation between the head entity and the tail entity in the given sentence.}\\
    \textbf{\textcolor{blue}{Relation:}} place served by transport hub; mountain range; religion; participating team; contains administrative territorial entity; head of government; country of citizenship; original network; heritage designation; performer; participant of; position held; has part; location of formation; located on terrain feature; architect; country of origin; publisher; director; father; developer; military branch; mouth of the watercourse; nominated for; movement; successful candidate; followed by; manufacturer; instance of; after a work by; member of political party; licensed to broadcast to; headquarters location; sibling; instrument; country; occupation; residence; work location; subsidiary; participant; operator; characters; occupant; genre; operating system; owned by; platform; tributary; winner; said to be the same as; composer; league; record label; distributor; screenwriter; sports season of league or competition; taxon rank; location; field of work; language of work or name; applies to jurisdiction; notable work; located in the administrative territorial entity;\\

    \textbf{\textcolor{blue}{Merpati}} flight 106 departed Jakarta ( CGK ) on a domestic flight to [Tanjung Pandan TAIL ENTITY] ( [TJQ HEAD ENTITY] ) . Relation: place served by transport hub\\
    \textbf{\textcolor{blue}{It is}} approximately 8 km away from [Mount Korbu HEAD ENTITY] , the tallest mountain of the [Titiwangsa Mountains TAIL ENTITY] . Relation: mountain range\\
    ...\\
    \textcolor{violet}{[unlabeled data} [[head entity] HEAD ENTITY]  [[tail entity] TAIL ENTITY]] Relation: \textcolor{orange}{[label]}\\
    }}
    };
    \end{tikzpicture}
    \caption{Prompt for PGDA3 used for FewRel Experiemtns.}
    \label{fig:prompt3forfewrel}
\end{figure}
\newpage
\begin{figure}[ht!]
    \centering
    \begin{tikzpicture}
    \node[draw, rounded corners] {
    \resizebox{0.9\linewidth}{!}{    
    \adjustbox{minipage=[r][49em][b]{0.46\textwidth},scale={0.7}}{
    \textbf{Identify the relation between the head entity and the tail entity in the given sentence.}\\
    \textbf{\textcolor{blue}{Relation:}} place served by transport hub; mountain range; religion; participating team; contains administrative territorial entity; head of government; country of citizenship; original network; heritage designation; performer; participant of; position held; has part; location of formation; located on terrain feature; architect; country of origin; publisher; director; father; developer; military branch; mouth of the watercourse; nominated for; movement; successful candidate; followed by; manufacturer; instance of; after a work by; member of political party; licensed to broadcast to; headquarters location; sibling; instrument; country; occupation; residence; work location; subsidiary; participant; operator; characters; occupant; genre; operating system; owned by; platform; tributary; winner; said to be the same as; composer; league; record label; distributor; screenwriter; sports season of league or competition; taxon rank; location; field of work; language of work or name; applies to jurisdiction; notable work; located in the administrative territorial entity;\\
    
    \textbf{\textcolor{blue}{MMerpati}} flight 106 departed Jakarta ( CGK ) on a domestic flight to Tanjung Pandan ( TJQ ) . <head> TJQ <tail> Tanjung Pandan <relation> place served by transport hub\\
    \textbf{\textcolor{blue}{It is}} approximately 8 km away from Mount Korbu , the tallest mountain of the Titiwangsa Mountains . <head> Mount Korbu <tail> Titiwangsa Mountains <relation> mountain range\\
    ...\\
    \textcolor{violet}{[unlabeled data]} <head> [head entity] <tail> [tail entity] <relation>: \textcolor{orange}{[label]}\\
    }}
    };
    \end{tikzpicture}
    \caption{Prompt for PGDA4 used for FewRel Experiemtns.}
    \label{fig:prompt4forfewrel}
\end{figure}
\newpage
\begin{figure}[ht!]
    \centering
    \begin{tikzpicture}
    \node[draw, rounded corners] {
    \resizebox{0.9\linewidth}{!}{    
    \adjustbox{minipage=[r][26em][b]{0.46\textwidth},scale={0.7}}{
    \textbf{\textcolor{blue}{Relation:}} place served by transport hub\\
    \textbf{\textcolor{blue}{Relation Definition:}} territorial entity or entities served by this transport hub (airport, train station, etc.)\\
    \textbf{\textcolor{blue}{Relation:}} mountain range\\
    \textbf{\textcolor{blue}{Relation Definition:}} range or subrange to which the geographical item belongs\\
    ...\\
    \textbf{Identify the relation between the head entity and the tail entity in the given sentence.}\\
    Sentence: Merpati flight 106 departed Jakarta ( CGK ) on a domestic flight to Tanjung Pandan ( TJQ ) .
    Head Entity: TJQ; Tail Entity: Tanjung Pandan\\
    \textbf{\textcolor{blue}{Relation:}} place served by transport hub\\
    \textbf{\textcolor{blue}{Sentence1:}} \textcolor{violet}{[unlabeled data]}\\
    \textbf{\textcolor{blue}{Head Entity1:}} [head entity]; Tail Entity1:[tail entity]\\
    \textbf{\textcolor{blue}{Relation:}} \textcolor{orange}{[label]}\\
    }}
    };
    \end{tikzpicture}
    \caption{Prompt for PGDA5 used for FewRel Experiemtns.}
    \label{fig:prompt5forfewrel}
\end{figure}

\subsection{Discussion on Annotation of More Complex Tasks}
\label{sec:complex_tasks}
The primary aim of this paper is to provide assistance to the majority of individuals and small and medium-sized enterprises (SMEs) engaged in data annotation. To achieve this objective, we have specifically chosen the most frequently employed tasks that possess broad downstream applications. In instances where more intricate tasks are encountered, previous research endeavors \cite{drozdov2023compositional,liu2023comprehensive, qin2023chatgpt} have demonstrated the remarkable capabilities of LLMs in complex NLP tasks, encompassing reasoning, semantic parsing, and dialogues, even in situations with minimal or no training data. Furthermore, there is a recent inclination to utilize LLMs for generating self-instruct data, exemplified by projects such as Alpaca\footnote{https://crfm.stanford.edu/2023/03/13/alpaca.html} and Vicuna\footnote{https://lmsys.org/blog/2023-03-30-vicuna/}. Consequently, the utilization of large language models for data annotation represents a nascent field with ample room for further research and enhancement.

\newpage

\newpage
\onecolumn
\subsection{Domain-Specific Knowledge of GPT-3}
\label{sec:specific_domain}

\begin{figure*}[h!]
    \centering
    \includegraphics[width=0.8\textwidth]{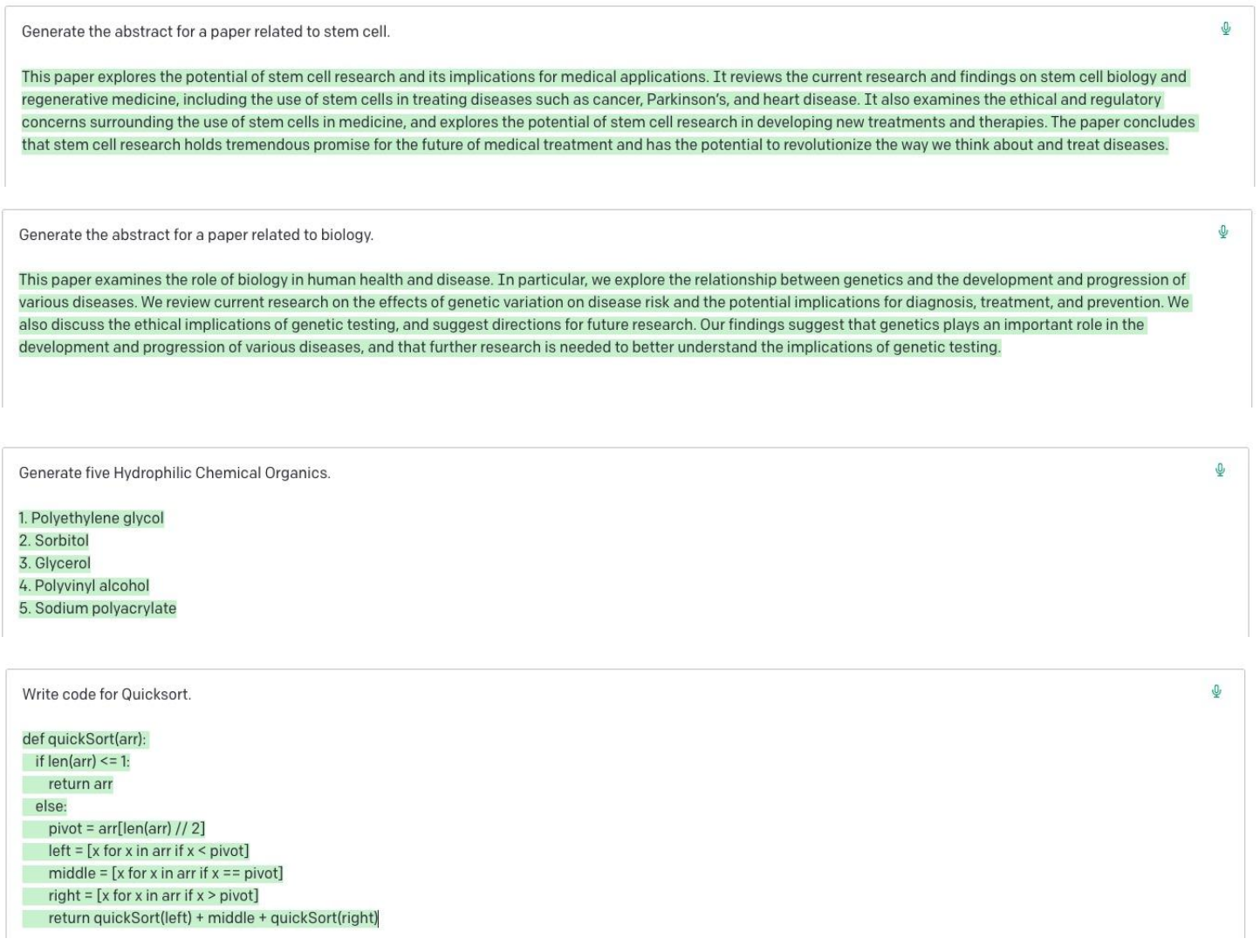}
    \caption{
    \small{Examples showing that GPT-3 has memorized a large amount of domain-specific knowledge.
    }
    \label{fig:domain_example_app}
    }
\end{figure*}

Figure~\ref{fig:domain_example_app} shows that GPT-3 has memorized a large amount of domain-specific knowledge.

\subsection{Results for SST2 under zero-shot and 2-shot settings}
\label{sec:sst2_2shot}
\begin{table*}[h!]
\centering
\scalebox{0.7}{
\begin{tabular}{lccccc}
\toprule

\textbf{Settings} & \textbf{Approach}              & \textbf{\begin{tabular}[c]{@{}c@{}}Number of Samples \\ Annotated / Generated\end{tabular}} & \textbf{\begin{tabular}[c]{@{}c@{}}Cost \\ (USD)\end{tabular}} & \textbf{\begin{tabular}[c]{@{}c@{}}Time \\ (Mins)\end{tabular}} & \textbf{Results} \\
\midrule
\multirow{6}{*}{Zero-shot} & \multirow{2}{*}{PGDA} & 3000                                    & 1.82       & 14\dag & 86.11          \\
                           &                       & 6000                                    & 3.65       & 27\dag & \textbf{87.31} \\
                           & \multirow{2}{*}{PGDG} & 3000                                    & 0.8        & 4\dag  & 78.25          \\
                           &                       & 6000                                    & 1.61       & 8\dag   & 80.15          \\
                           & \multirow{2}{*}{DADG} & 3000                                    & 3.10       & 13\dag & 73.53          \\
                           &                       & 6000                                    & 6.21       & 25\dag  & 76.66          \\
\midrule
\multirow{6}{*}{2-shot}    & \multirow{2}{*}{PGDA} & 3000                                    & 3.18       & 16 & 85.89          \\
                           &                       & 6000                                    & 6.36       & 32\dag  & \textbf{89.07} \\
                           & \multirow{2}{*}{PGDG} & 3000                                    & 0.97       & 4\dag   & 79.57          \\
                           &                       & 6000                                    & 1.94       & 9\dag & 79.24          \\
                           & \multirow{2}{*}{DADG} & 3000                                    & 3.68       & 15\dag & 75.34          \\
                           &                       & 6000                                    & 7.38       & 29\dag & 77.32      \\ 
\bottomrule
\end{tabular}
 }
\caption{Costs, time spending, and results of SST2 under zero-shot and 2-shot settings. \dag means multiprocessing (5 processes) is enabled. Time for manual labeling excludes the time spent on instruction preparation and training.}
\label{tb:sst_result_0shot_2shot}

\newpage


\end{table*}

\newpage
\subsection{Case Study of Multilingual Data Annotation}
\label{sec:multilingual}
\begin{figure*}[h!]
    \centering
    \includegraphics[scale=1.0]{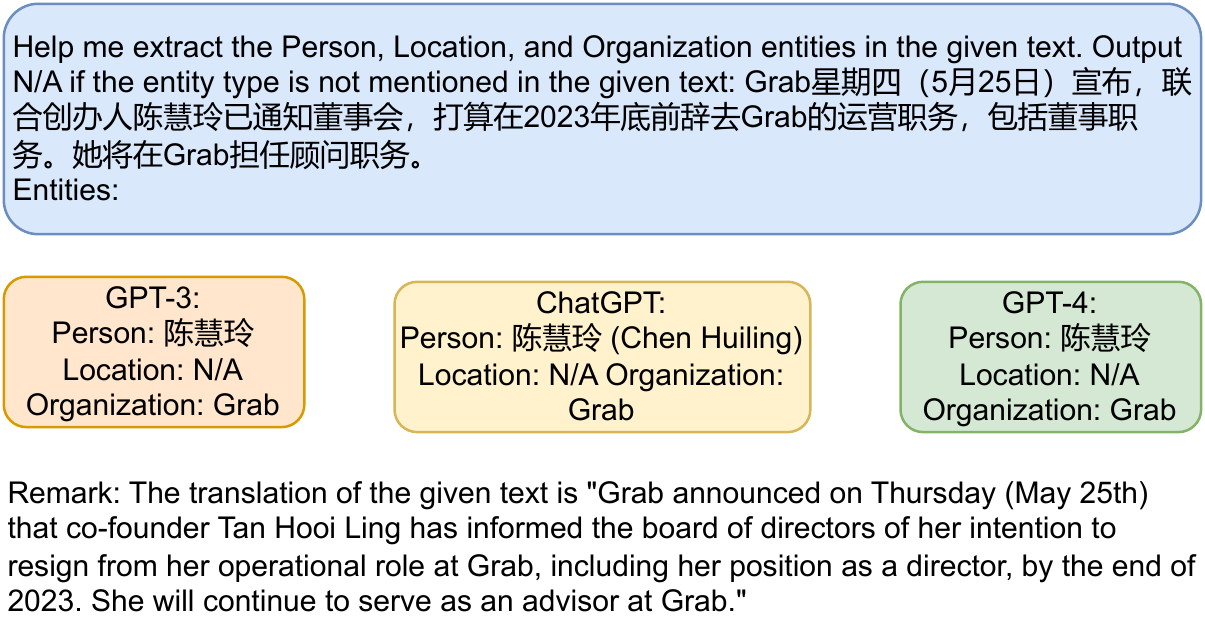}
    \caption{Illustrations of Annotating Chinese NER using GPT-3, ChatGPT and GPT-4.}
    \label{fig:case} 
\end{figure*}

\begin{figure*}[h!]
    \centering
    \includegraphics[scale=1.0]{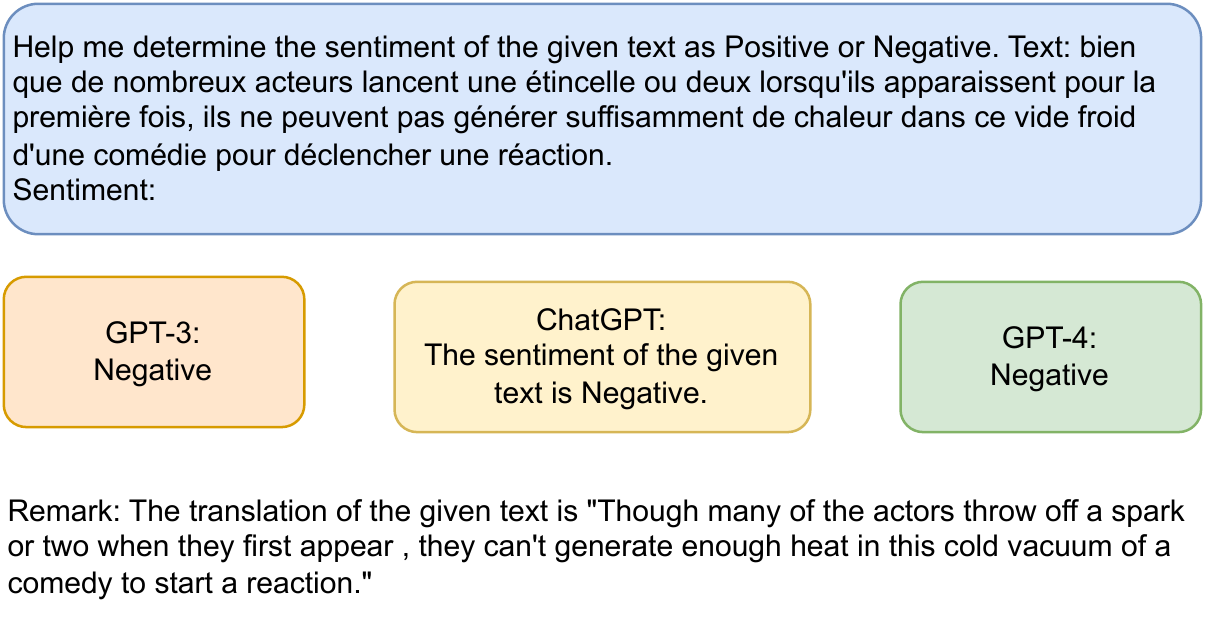}
    \caption{Illustrations of Annotating French Text Classification Data using GPT-3, ChatGPT and GPT-4.}
    \label{fig:case2} 
\end{figure*}

Figure~\ref{fig:case} and \ref{fig:case2} shows that GPT-3, ChatGPT and GPT-4 can be used to annotate data in non-English languages.

